\documentclass[pdflatex,sn-mathphys,iicol]{sn-jnl}

\usepackage{pdfpages}
\usepackage{graphicx}
\usepackage{float}
\usepackage{amssymb,amsfonts,amsmath,amsthm}
\usepackage{txfonts}

\usepackage{breqn}
\usepackage{algorithm,algpseudocode}
\usepackage{enumitem}

\begin{document}

\title[Image captioning]{Image Captioning based on Feature Refinement and Reflective Decoding}


\author{\fnm{Ghadah} \sur{Alabduljabbar}}\email{439204033@student.ksu.edu.sa}

\author{\fnm{Hafida} \sur{Benhidour}}\email{hbenhidour@ksu.edu.sa}

\author{\fnm{Said} \sur{Kerrache}}\email{skerrache@ksu.edu.sa}

\affil{\orgdiv{Department of Computer Science}, \orgname{King Saud University}, \orgaddress{\city{Riyadh}, \postcode{11543}, \country{Saudi Arabia}}}

%


\abstract{Image captioning is the process of automatically generating a description of an image in natural language. Image captioning is one of the significant challenges in image understanding since it requires not only recognizing salient objects in the image but also their attributes and the way they interact. The system must then generate a syntactically and semantically correct caption that describes the image content in natural language. With the significant progress in deep learning models and their ability to effectively encode large sets of images and generate correct sentences, several neural-based captioning approaches have been proposed recently, each trying to achieve better accuracy and caption quality. 
This paper introduces an encoder-decoder-based image captioning system in which the encoder extracts spatial features from the image using ResNet-101. This stage is followed by a refining model, which uses an attention-on-attention mechanism to extract the visual features of the target image objects, then determine their interactions. The decoder consists of an attention-based recurrent module and a reflective attention module, which collaboratively apply attention to the visual and textual features to enhance the decoder's ability to model long-term sequential dependencies. Extensive experiments performed on Flickr30K, show the effectiveness of the proposed approach and the high quality of the generated captions.}




\maketitle

\section{Introduction}\label{sec:introduction}

Visual information is pervasive in our daily life and has arguably become the primary communication medium in the digital age. Images come from various sources, such as social networks, advertisements, news, and digital documents. A small portion of these images have associated description that helps understand, organize and search their content, but the vast majority does not. Image captioning using machine learning is the process of automatic generation of a natural language description for an input image. In order to automatically generate an appropriate caption, the system must detect the various objects in the image and then describe them in natural language. 

Image captioning systems have numerous practical applications, such as helping visually impaired individuals better understand the content of images, hence improving their user experience. They can also be integrated into guidance devices that help improve the life quality of the visually impaired. Automatic image indexing can benefit from image captioning since it generates more sophisticated and semantically rich captions than image classification and tagging. Automatic image indexing is essential in many areas like education, medicine, e-commerce, digital libraries, military, and web searching. Furthermore, image captioning can improve the performance of content-based image retrieval applications which search for a specific image based on its content. Social media platforms can use image captioning to generate accurate descriptions of images posted by users and provide better personal recommendation and content filtering. More generally, all domains where humans need to describe or interact with images in any form will benefit from a system capable of automatically generating textual image captions.

Even though effective algorithms for recognizing objects in videos and images have been developed in recent decades \cite{torralba2003context}, describing image visual scenes and features in well-formed sentences remains a very challenging task. Such systems must extract semantic concepts and visual features and transform them into correctly formed sentences in natural language \cite{vinyals2016show}. Generating accurate textual descriptions of images chiefly depends on extracting rich image global visual features. Early work in image captioning mainly used template-based and retrieval-based methods. Both types tried to solve image captioning problems by using training captions or hard-coded language rules. However, these techniques have significant limitations that affect the generated caption quality when used with large, complex, and diverse data that admit various semantic interpretations. With the considerable progress in research on neural networks and their proven capability for handling complex learning tasks, neural network-based approaches have become the dominant approach in image captioning. Deep learning techniques use training data to learn complex and rich image features allowing them to handle large, complex, and diverse datasets. Despite this progress, the challenge remains for researchers to generate better quality captions using more sophisticated visual and language network architectures.

This work proposes an encoder-decoder image captioning system that uses a refining module on the spatial features extracted from ResNet-101 in the encoder and a reflective decoding module in the decoder. Unlike previous approaches that attempt to improve the performance by focusing on improving visual features alone, our model uses the extracted visual features and an attention-on-attention mechanism \cite{huang2019attention}  to model the relationships among objects in the target image. The decoder uses a reflective decoding module\cite{ke2019reflective} that applies two types of attention to the visual and textual information to focus on the most salient objects in the image and obtain more comprehensive information to generate higher-quality captions. The proposed approach is implemented and evaluated using Flickr30k, a benchmark image captioning dataset. The experimental results analysis demonstrate the proposed approach's effectiveness. 

This paper is organized as follows. First, we survey the existing literature on image captioning in Section \ref{sec:related-work} and present a comparative study of related work. The proposed approach is presented in detail in Section \ref{sec:proposed}. Next, we provide a detailed experimental evaluation of the proposed architectures. We finally conclude this paper in Section \ref{sec:conclusion} and give an overview of future work.

\section{Related Work}
\label{sec:related-work}
Automatic generation of image description is known within the research community as image captioning. It is the process of automatically generating a well-formed caption that describes the important objects in an image, including their location, type, properties, and relationships. Following \cite{hossain2019comprehensive, bai2018survey}, we can broadly classify the existing image captioning approaches into two main categories: early work and neural network-based techniques. Early work approaches mainly used template-based and retrieval-based methods for generating captions. However, following the significant progress in neural networks research in the previous decade, neural-based approaches for image captioning became the dominant trend in the field. Deep learning methods achieved a remarkable improvement in performance and demonstrated excellent capabilities for handling the challenges and complexities of the image captioning task. In what follows, we will present both categories of captioning approaches and highlight their weaknesses and strengths. 

\subsection{Early work on image captioning}
Early image captioning systems mainly used retrieval-based and template-based approaches. Authors in \cite{farhadi2010every} and \cite{hodosh2013framing} used retrieval-based image captioning by mapping the target image to a space of images and captions retrieved from the dataset. The authors in \cite{farhadi2010every} used Markov Random Field for the mapping and Lin similarity measure to find the best caption. In contrast, \cite{hodosh2013framing} used Kernel Canonical Correlation Analysis (KCCA) technique for the mapping and cosine similarity measure for captions ranking. Retrieval-based methods can generate syntactically correct captions. However, since these generated captions are obtained by reusing and composing existing captions, they generally are semantically inaccurate. In addition to the lack of large datasets available in recent years like COCO by Microsoft \cite{lin2014microsoft} and Conceptual Captions by Google \cite{sharma2018conceptual}, using relatively small datasets negatively influences the quality of the generated captions.

On the other hand, \cite{yang2011corpus} and \cite{mitchell2012midge} used template-based image captioning by detecting features from the target image and then using templates to generate captions. The authors in \cite{yang2011corpus} used Hidden Markov Model inference (HMM) to fill the template structure. Unlike \cite{yang2011corpus} or any previous approaches that used a fixed template \cite{mitchell2012midge} used a computer vision detector system to produce a syntactic tree that will summarize the extracted image features to generate caption, which results in better quality captions. Even though the template-based approach may generate better and more relevant descriptions than the retrieval-based approach, it still has limitations regarding caption creativity, coverage, and complexity. Furthermore, template-based captions lack naturality compared to human written captions due to using predefined templates.

Early work captioning systems, whether retrieval-based or template-based, suffer obvious limitations; They are not flexible enough to generate expressive and meaningful captions. With the significant progress in neural networks research, especially in deep neural networks, many recent works tackled these limitations by adopting neural networks in automatic image captioning using various methods and architectures.

\subsection{Neural network-based image captioning}
The first attempt to use neural networks to solve image captioning problems was in \cite{kiros2014multimodal}. The authors replaced templates, rules, and constraints with Convolutional Neural Networks (CNN) \cite{lecun1998gradient} and Multimodal Neural Language Models (MNLM) to generate better quality captions. They then extended their work in \cite{kiros2014unifying} to introduce an encoder-decoder framework that uses deep CNN as a visual model and Long Short-Term Memory (LSTM) as a language model. The Neural Image Caption Generator (NIC) model proposed in \cite{vinyals2015show} is similar to \cite{kiros2014unifying} but is trained using maximum likelihood estimation to maximizing the likelihood of observing the ground truth captions. These encoder-decoder frameworks significantly improved the quality of the generated captions grammatically and syntactically. However, their performance depends primarily on the quality of global and high-level features extracted from the target image using CNN and the effectiveness of captions representation by neural language models.

Instead of an end-to-end encoder-decoder image captioning framework, \cite{fang2015captions} and \cite{tran2016rich} used compositional architecture-based approaches that consist of a visual model, a language model, and a multimodal similarity model. The authors of \cite{fang2015captions} used a visual model that focused on image subregions, unlike previous models that operate on the whole image, a Maximum Entropy (ME) language model, and a Deep Multimodal Similarity Model (DMSM) to choose the best caption for the target image. Based on the architecture proposed in, \cite{fang2015captions}  \cite{tran2016rich} built a vision model that specializes in generating captions for open domain images. However, instead of extracting visual features using CNN, they used a deep Residual Network (ResNet) with entity detection to identify celebrities and landmarks, resulting in improved performance on in- and out-of-domain datasets. However, even though using multiple models can improve the performance and make the image captioning system robust and more powerful, it also increases the system's overall complexity. 

Compared to methods that focus on the global features in the whole image, the attention mechanism allows focusing on the most salient objects of the target image before forming a caption. The first to use such a technique in image captioning were \cite{xu2015show}. They learn the hidden alignments from scratch instead of detecting image objects like other methods to focus on the important parts of the target image. On the other hand, \cite{lu2017knowing} used a visual sentinel that considers the none visual words in addition to the ones predicted by the language model. Authors in \cite{pedersoli2017areas} used a spatial transformer network to define the attention areas in the target image without using bounding boxes in the training process. Even though \cite{pedersoli2017areas} was faster, \cite{lu2017knowing} generates better caption quality. An attention-based mechanism helps improve the quality and relevance of the generated captions and reduces computational complexity since it skips unimportant details in the image. However, it has limitations like neglecting some details in the target image that might be important for caption generation and the difficulty in formulating image captioning as an end-to-end process. 

Previously discussed image captioning approaches used either top-down mechanism \cite{donahue2015long, karpathy2015deep} or bottom-up mechanism \cite{farhadi2010every, kulkarni2013babytalk}. In top-down approaches, one starts by extracting visual image features and then converts them to words. In contrast, bottom-up approaches start by extracting visual image concepts like attributes, regions, and objects and then combining them with words that describe them. Both mechanisms have limitations, such as neglecting some details in the target image that might be important for caption generation in top-down approaches or the problem of formulating an end-to-end process in bottom-up methods. To overcome these limitations, You et al. \cite{you2016image} combined top-down and bottom-up approaches to selectively focus on the image's semantically important concepts and work at any resolution and in any region of the target image. In \cite{wu2017image}, the authors focus on generating semantically sound captions by extracting the high-level semantic concepts from the target image using an attribute prediction layer. Using a semantic concept-based approach can generate semantically rich image captions. However, focusing on some parts of the image might cause the system to miss some small details in the target image that might be helpful in the caption generation process.

\subsection{Discussion}	

\begin{table*}[!t] 
	\centering
	\caption{Literature approaches comparison (MSCOCO Dataset)}
	\label{table:1}
	\footnotesize
	\begin{tabular}{ l l l l l l l l }
		\toprule
		Ref & Architecture & Encoder & Decoder & Modeling Space & Attention Mechanism  & METEOR & CIDEr \\ 
		\midrule
		\cite{vinyals2015show} & Encoder-Decoder & GoogLeNet & LSTM & Visual & No & 0.237 & 0.855 \\ 
		\hline
		\cite{fang2015captions} & Compositional & AlexNet, VGGNet & MELM & Visual & No & 0.236 & - \\ 
		\hline
		\cite{xu2015show} & Encoder-Decoder & AlexNet & LSTM & Visual & Yes & 0.239 & - \\ 
		\hline
		\cite{you2016image} & Encoder-Decoder & GoogLeNet & RNN & Visual & Yes & 0.243 & 0.958 \\ 
		\hline
		\cite{lu2017knowing} & Encoder-Decoder & ResNet & LSTM & Visual & Yes & \textbf{0.266} & \textbf{1.085} \\
		\hline
		\cite{kiros2014multimodal} & Encoder-Decoder  & AlexNet & LBL & Multimodal & No & 0.203 & - \\
		\hline
		\cite{wu2017image} & Encoder-Decoder & VGGNet & LSTM & Visual & Yes & 0.26 & 0.94 \\ 
		\hline
		\cite{pedersoli2017areas} & Encoder-Decoder & VGGNet & RNN & Visual & Yes & 0.252 & 0.981 \\ 
		\hline
		\cite{kiros2014unifying} & Encoder-Decoder & AlexNet, VGGNet & LSTM, SC-NLM & Multimodal & No & - & - \\
		\hline
		\cite{tran2016rich} & Compositional & ResNet & MELM & Visual & No & - & - \\
		\bottomrule
	\end{tabular}
\end{table*}

Table~\ref{table:1} compares the research works previously mentioned in the literature in terms of architecture, visual model, language model, modeling space, and attention mechanisms. We can observe from the table that most approaches adopt the encode-decoder architecture since it produces high performance with minimum complexity and considerable robustness. In contrast, only a few of them \cite{fang2015captions}, \cite{tran2016rich} adopt the compositional architecture. 

In terms of the visual model used to encode the target image, there are five convolutional neural networks that are commonly used which are VGGNet \cite{simonyan2014very}, ResNet \cite{he2016deep}, AlexNet \cite{krizhevsky2012imagenet}, DenseNet and GoogleNet. ResNet performs the best compared to others, but VGGNet is still the most popular among researchers even though it produces the second highest result. This choice is due to the fact that VGGNet is easier to implement, and recent research efforts focused more on simplicity and speed.

Regarding the language model, Long-Short-Term Memory (LSTM) \cite{hochreiter1997long}, which is mainly used as a decoder, is the most popular language model. Researchers tried other models such as Recurrent Neural Networks (RNN) \cite{rumelhart1986learning}, Maximum Entropy Language Model (MELM), Log-Bilinear Language Model (LBL), and Structure-Content Neural Language Model (SC-NLM). However, LSTM remains the model of choice because of the quality of its generated captions. 

Table~\ref{table:1} also shows that most approaches model the image in a visual space which offers a faster computational speed, but some approaches learn from combining visual and language space.

Finally, using an attention mechanism enables the approaches to dynamically focus on various parts of the image while the captions are being generated or selectively focus on a group of semantic concepts  \cite{you2016image}, \cite{wu2017image}. As a result, attention mechanisms help generate high-quality and relevant captions. 

\section{Proposed Approach}
\label{sec:proposed}
We propose an end-to-end image captioning system based on an encoder-decoder framework with attention mechanisms. The encoder-decoder is the most used architecture for image captioning systems, and an end-to-end approach means both the encoder and the decoder are used in a tightly coupled manner. Hence the model is trained as a whole, offering good performance with less engineering effort. In contrast to other approaches that use spatial or global features and focus on the visual features alone, our approach uses the refined visual features extracted using ResNet-101. We then apply two attention mechanisms in the decoder, the first to the visual features to focus on the image's salient parts and the second to the textual features to generate captions with more detailed information.


\begin{figure*}[!t]
	\centering 
	\includegraphics[width=0.7\linewidth]{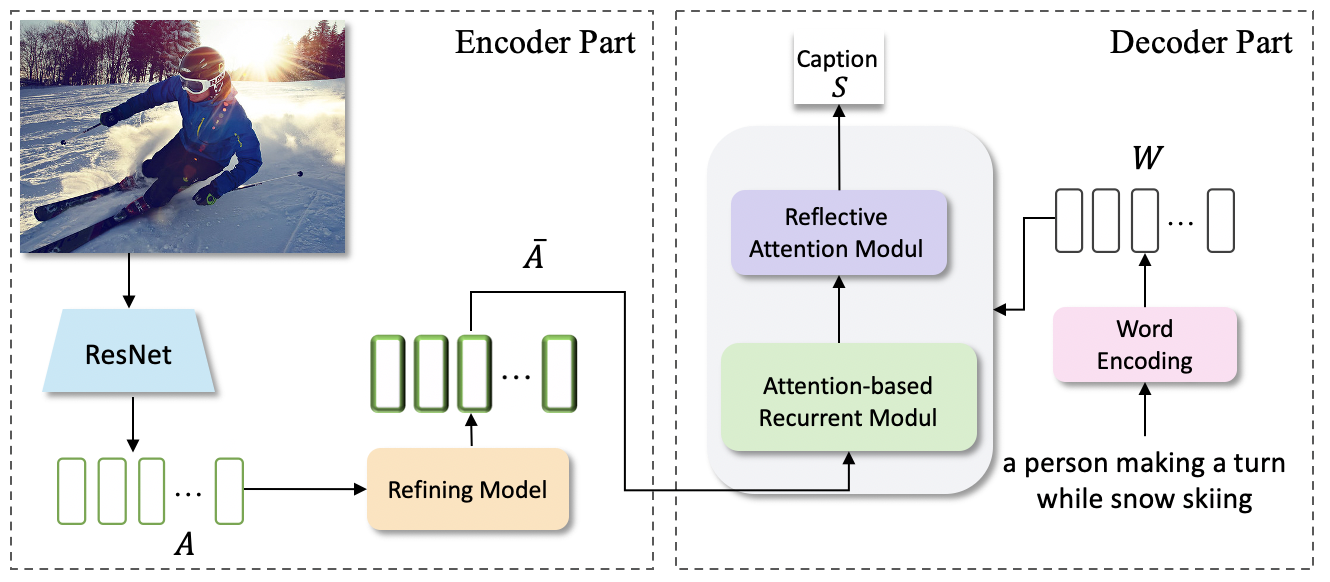}
	\caption{Overview of the proposed architecture. A refining model is used to smooth the representations of the extracted object features.}
	\label{fig:PA2}
\end{figure*}

High-level views of the proposed architecture is shown in Figure~\ref{fig:PA2}. 
The encoder of the proposed architecture consists of ResNet-101 \cite{he2016deep} and a refining model \cite{huang2019attention}, which smooths the representations of the extracted object features by modeling relationships among them. The extracted image features are then fed to the decoder.

The decoder part consists of two modules. The first is an attention-based recurrent module \cite{anderson2018bottom} used to attend selectively to the visual features extracted by the encoder. The second module is a reflective attention module (RefAM) \cite{ke2019reflective} that enhances the caption quality by capturing more historical coherence and comprehensive information. The second module is designed on top of the first one for further improvement in caption quality.

The subsections below describe each component of the proposed architectures in more detail.

\subsection{The Encoder}

As shown in Figure~\ref{fig:PA2}, given an image $M$, we first use a pre-trained ResNet-101 to extract a set of visual feature vectors (spatial features) from the last convolutional layer of the CNN. The extracted features are denoted by $A = \{a_1,a_2, \ldots, a_k\}, a_i \in \mathbb{R}^{D}$, where $k$ is the number of the extracted vectors. Inspired by \cite{lu2017knowing}, we also extract the global image feature vector $A_g$, which are features taken from the last fully connected layer of CNN.	

ResNet-101 is pre-trained on more than a million images from the ImageNet database \cite{russakovsky2015imagenet} to learn rich visual feature representations for various images. It can classify objects in images into 1000 categories. The spatial features are extracted from the last convolutional layer, whereas the global features are taken from the last fully connected layer. The ResNet-101 network takes a raw image $M$ as an input, then encodes it into $k$ vectors of size $D$. Each vector represents the visual features extracted from different locations of the image $M$:
\begin{equation}  
	A = {CNN}(M) = \{ a_1, a_2, \ldots , a_k\}.
\end{equation}
The residual network stacks layers to fit a residual mapping, resulting in greatly enriched feature vectors. Each layer in the network implements the following transformation:
\begin{equation}  
	A_{out} = \mathcal{F}(A_{in}, \{W_i\}) + A_{in},
\end{equation}
where:
\begin{itemize}
	\item $ A_{in} $ is the input vector of the current layer.
	\item $ A_{out} $ is the output vector of the current layer.
	\item $ \mathcal{F} $ denotes the residual mapping function of convolutional layers.
	\item $ W_i $ is the parameters set.
\end{itemize} 


\begin{figure}[!t] 
	\centering 
	\includegraphics[width=\linewidth]{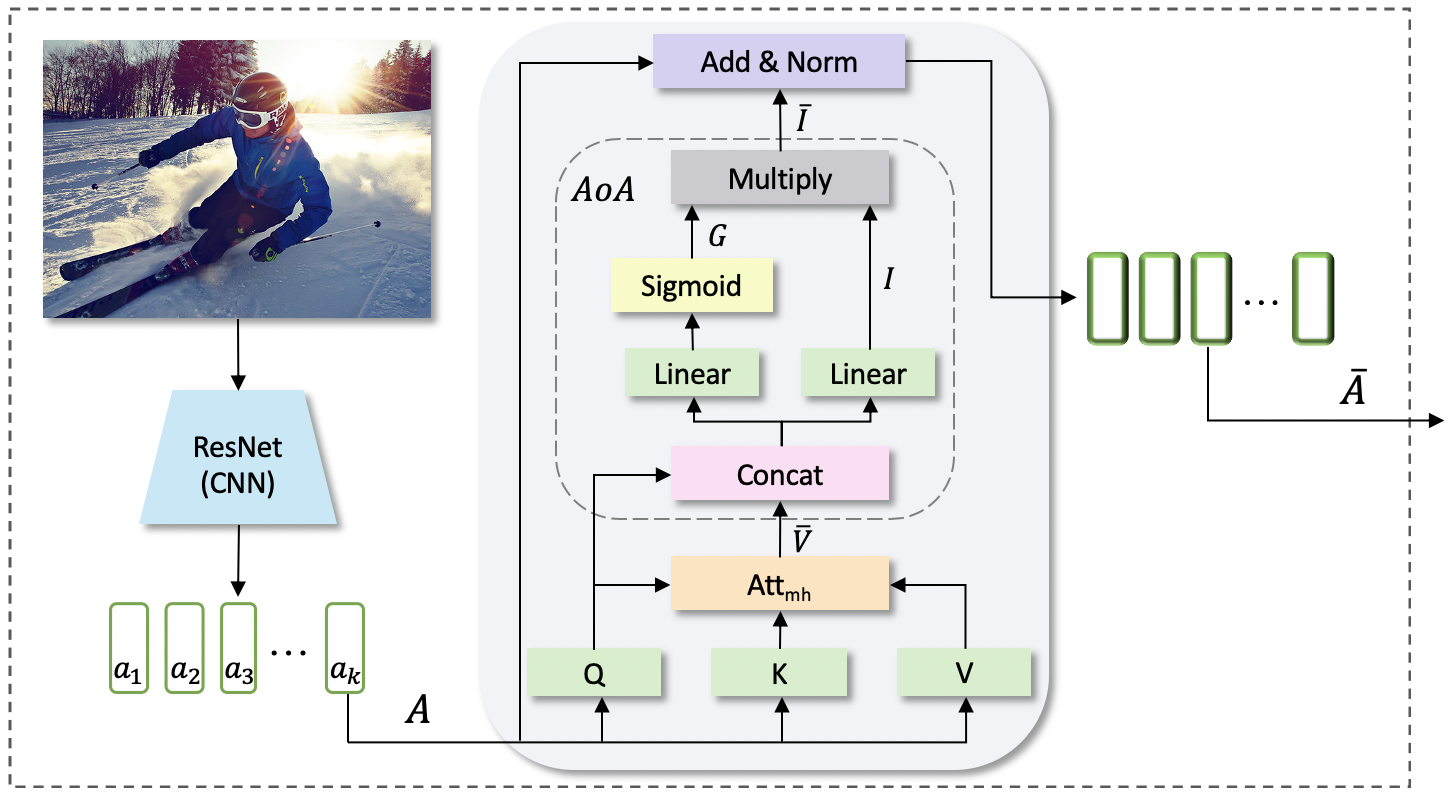}
	\caption{The proposed encoder architecture.}
	\label{fig:PAEn}
\end{figure}

Figure~\ref{fig:PAEn} gives a detailed view of the encoder. The features $A = \{ a_1, a_2, \ldots, a_k\}$ extracted by ResNet-101 are fed into the refining network to refine their representation by modeling the relationships among them. The refining model consists of three individual linear functions $Q$ (Query), $K$, and $V$ (Key/Value pair) that perform linear projections of the features vector $A$. The output is then fed into a multi-head attention function ($Att_{mh}$) \cite{vaswani2017attention}, followed by an attention-on-attention ($AoA$) model \cite{huang2019attention}, a residual connection \cite{he2016deep}, and finally layer normalization \cite{ba2016layer}.

The multi-head attention module $Att_{mh}$ divides each $Q$, $K$, $V$ into slices along the channel dimension, then applies the self attention function $f_{Att_{dot}}$ to every slice $Q_i$, $K_i$, $V_i$. Afterward, the result of each slice is concatenated to produce the final attended vector $\bar{V}$:
\begin{align}  
	f_{Att_{dot}}(Q_i,K_i,V_i) &= Softmax\left(\dfrac{Q_iK_i^T}{\sqrt{d}}\right)V_i,\\
	head_i &= f_{Att_{dot}}(Q_i,K_i,V_i), \\
	\bar{V} = f_{Att_{mh}}(Q,K,V) &= Concatenate(head_1, \ldots ,head_H).
\end{align}

The attention-on-attention model ($AoA$) measures how well the objects of the target image are related by measuring the relevance between the multi-head attention result $\bar{V}$ and the query $Q$. After concatenating  $\bar{V}$ and $Q$, the model generates the information vector $I$ and the attention gate $G$ conditioned on  $\bar{V}$ and the current context $Q$. To obtain the final attended information vector $\bar{I}$, the model uses element-wise multiplication between the attention gate $G$ and the information vector $I$:
\begin{align}  
	I &= W^I_q Q + W^I_v \bar{V} + b^I, \\
	G &= \sigma (W^G_q Q + W^G_v \bar{V} + b^G), \\
	\bar{I} &= AoA(\bar{V}, Q) = \sigma (G) \odot (I),
\end{align}
where:
\begin{itemize}
	\item $ W^I_q, W^I_v, W^G_q, W^G_v $ are linear transformation matrices $\in \mathbb{R}^{D\times D}$.
	\item $ D $ is the dimension of $Q$ and $V$.
	\item $ \sigma $ is the sigmoid activation function.
	\item $ \odot $ is element-wise multiplication.
\end{itemize} 

After refining the extracted features, the feature vector $A$ is updated to the refined features vector $\bar{A}$, without changing its dimension as follows:
\begin{equation}  
	\bar{A} = LayerNorm(A + \bar{I}),
\end{equation}
where $LayerNorm$ is the layer normalization proposed in \cite{ba2016layer}:
\begin{align}  
	\mu_i &= \frac{1}{m} \sum_{j=1}^{m} x_{ij}, \\
	\sigma^2_i &= \frac{1}{m} \sum_{j=1}^{m} (x_{ij} - \mu_i)^2, \\
	\hat{x}_{ij} &= \frac{x_{ij} - \mu_i}{\sqrt{\sigma^2_i + \epsilon}},
\end{align}
where $ x_{ij} $ is the $(i,j)$-th element of the input.

\subsection{The Decoder}

\begin{figure} [!t] 
	\centering 
	\includegraphics[width=0.9\linewidth]{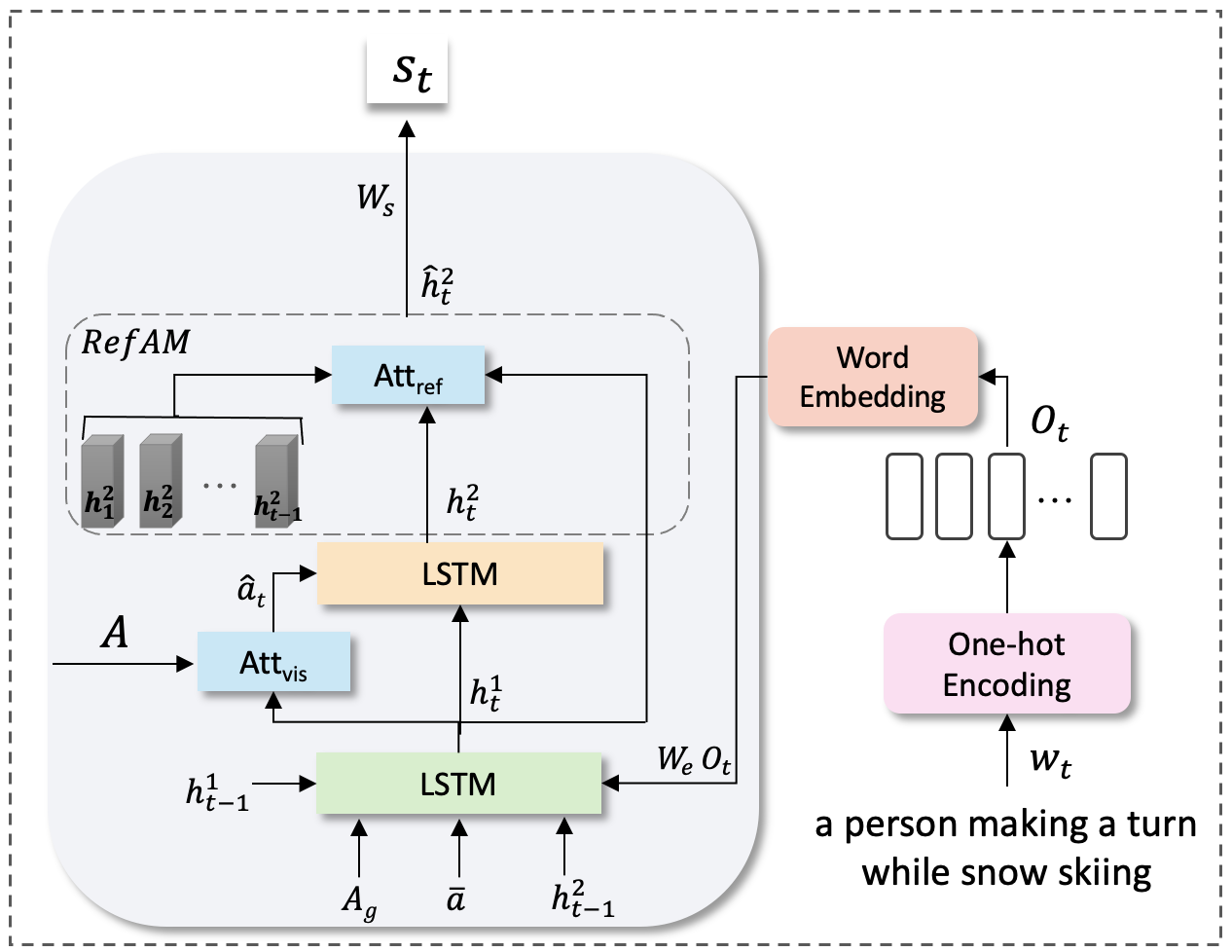}
	\caption{The proposed decoder architecture.}
	\label{fig:PADe}
\end{figure}

Given the set of spatial features vector $A= \{a_1,a_2, \ldots , a_k\}$ produced by the encoder, global image feature vector $A_g$, and the embedding word vector $W$, the decoder aims to generate a caption $S$ of $n$ words, where $S = \{s_1, s_2, \ldots , s_n\}$. As shown in Figure~\ref{fig:PADe}, the decoder consists of two modules, an attention-based recurrent module, and a reflective attention module. The first module accomplishes the main function of the decoder. It contains two LSTM layers and one layer of top-down visual attention ($Att_{vis}$), which determines the distribution of attention over all detected regions of the target image. The input $x^1_t$ of the first LSTM layer at time step $t$ is obtained by concatenating four components: 
\begin{enumerate}
	\item The global image features $A_g$.
	\item The image contextual information $\bar{a}$  obtained by mean pooling of the image features $A$.
	\item The word embedding $w_t$.
	\item The previous output of the second LSTM layer $h^2_{t-1}$. 
\end{enumerate}
The units of the first LSTM layer  are updated as follows:
\begin{align}  
	\bar{a} &= \frac{1}{k} \sum_{i=1}^{k} a_i,\\
	x^1_t &= [ A_g, \bar{a}, W_e O_t, h^2_{t-1}],\\
	h^1_t &= LSTM(x^1_t,h^1_{t-1}),
\end{align}
where:
\begin{itemize}
	\item $ W_e $ is the word embedding matrix for the one-hot encoding vector $O_t$ ($O_t \in \mathbb{R}^{D_o}$) of the input word $w_t$, $W_e \in \mathbb{R}^{E\times D_o}$.
	\item $ E $ is the embedding size.
	\item $ D_o $ is the vocabulary size.
\end{itemize} 

Next is $Att_{vis}$ layer, the input is the generated state $h^1_t$, and the set of visual features $A = \{a_i\}^k_{i=1}$ obtained from the encoder, the $Att_{vis}$ calculates the weight $\alpha^{vis}_t$ of the normalized attention distribution over all target image regions as follow:

\begin{align}  
	\alpha^{vis}_{i,t} &= W^1_v \tanh \left(W^1_{rv} a_i + W^1_{hv} h^1_t \right),\\
	r^{vis}_t &= \{\alpha^{vis}_{i,t}\}^k_{i=1},\\
	\alpha^{vis}_t &= Softmax(r^{vis}_t),\\
	\hat{a}_t &= \sum_{i=1}^{k} \alpha^{vis}_{i,t} a_i,
\end{align}
where:
\begin{itemize}
	\item $ W^1_v $ denotes the learned embedding matrices $\in \mathbb{R}^{1\times D_v}$.
	\item $ W^1_{rv/hv} $ denotes the learned embedding matrices $W^1_{rv} \in \mathbb{R}^{D_v\times D} , W^1_{hv} \in \mathbb{R}^{D_v\times D_{h}}$.
	\item $ \alpha^{vis}_t $ is a vector of attention probability scores of visual each feature $a_i$ at time step $t$, $\alpha^{vis}_t \in \mathbb{R}^{1\times D}$
\end{itemize} 

The second layer of LSTM takes the $Att_{vis}$ attention result $ \hat{a}_t$ which defines the attended features, and the generated hidden state $h^1_t$ from the first LSTM. The updating formula for the second LSTM layer units are as follows:
\begin{align}  
	h^2_t &= LSTM(x^2_t,h^2_{t-1}),\\
	x^2_t &= [ \hat{a}_t, h^1_t],
\end{align}
where $ h^2_t $ is the LSTM hidden state $\in \mathbb{R}^{D_h}$.

\begin{figure} [!t] 
	\centering 
	\includegraphics[width=0.7\linewidth]{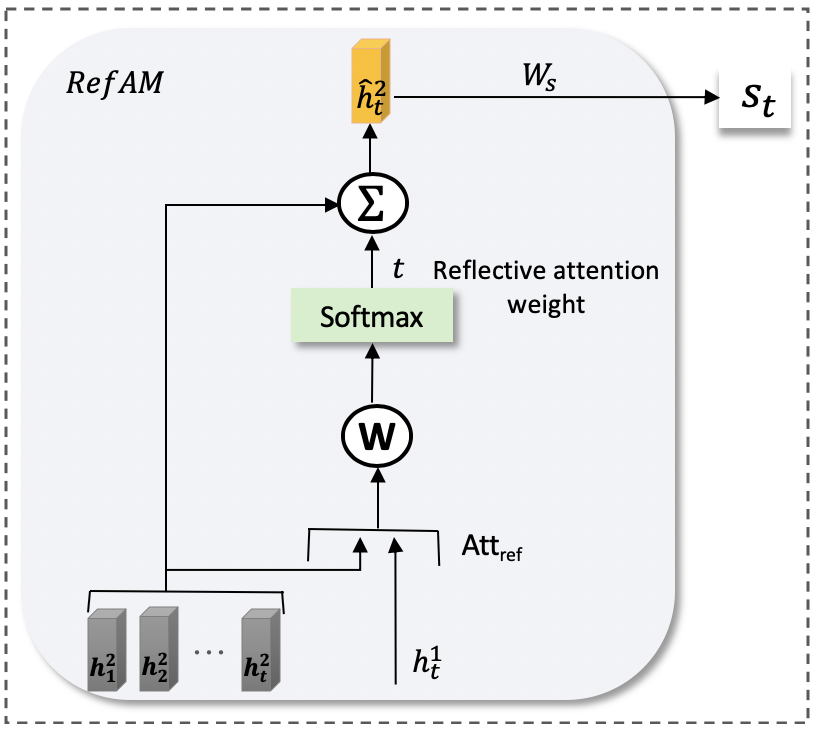}
	\caption{The reflective attention module architecture \cite{ke2019reflective}.}
	\label{fig:RefDe}
\end{figure}

The reflective attention module (RefAM) consists of a reflective attention layer $Att_{ref}$ that selectively attends to the past hidden states and the second LSTM  layer. As reflected in Figure~\ref{fig:RefDe}, the $Att_{ref}$ layer takes the current hidden state $h^2_t$ generated from the second LSTM layer and the set of all previous hidden states $\{ h^2_1, h^2_2, \ldots , h^2_{t-1} \}$ as an input to compute the normalized weight distribution $ \alpha^{ref}_t $ over all generated hidden states:
\begin{align}  
	\alpha^{ref}_{i,t} &= W^2_h \tanh(W^2_{h_2h} h^2_i + W^2_{h_1h} h^1_t),\\
	r^{ref}_t &= \{\alpha^{ref}_{i,t}\}^t_{i=1},\\
	\alpha^{ref}_t &= Softmax(r^{ref}_t),
\end{align}
where:
\begin{itemize}
	\item $ W^2_h $ is a trainable matrix in $\mathbb{R}^{1\times D_f}$.
	\item $ W^2_{h_{1} h} $ and $ W^2_{h_2 h} $ denote trainable matrices in $\mathbb{R}^{{D_f}\times D_h}$.
	\item $ \alpha^{ref}_t $ is the set of the generated attention probabilities for all hidden states, $\alpha^{ref}_t \in \mathbb{R}^{1\times D_h}$.
\end{itemize}

The weight $\alpha^{ref}_{i,t}$ measures the compatibility of the previously generated hidden state $h_i$ and the current hidden state to demonstrate the relevance between them. The reflective decoding then computes the attended hidden state $\hat{h}^2_t \in \mathbb{R}^{D_h}$, which is used to predict the next word $s_t$:
\begin{align}  
	\hat{h}^2_t &= \sum_{i=1}^{t} \alpha^{ref}_{i,t} h^2_i,\\
	p(s_t|s_{1:t-1}) &= Softmax(W_s \hat{h}^2_t + b_s),
\end{align}
where:
\begin{itemize}
	\item $ W_s $ are the weights $\in \mathbb{R}^{{D_o}\times D_h}$.
	\item $ b_s $ are the biases $\in \mathbb{R}^{D_o}$.
	\item $ D_o $ is the captioning vocabulary size.
\end{itemize}
Using the set of all generated hidden states  $\{h_i\}^t_{i=1}$ to calculate the next word $s_t$ and model the dependencies between each word pair explicitly at different time steps is expected to increase word precision and produce higher quality captions.

\subsection{Beam Search}
Once the model is trained, it can be used to generate new captions in the inference stage. After feeding the $ \textlangle$start$\textrangle $ token to the LSTM, the generation process continues until the $ \textlangle$end$\textrangle $ token is reached. Beam search is used to select the next token in the evaluation stage, enabling the model to choose the sequence with the best overall score among several candidate sequences conditioned on the input image. At each decoding step, the search algorithm considers a set of top $k$ sentences at time $t$ as possible candidates to generate a sentence at time $t+1$  ($k$ is the beam size). This process is repeated until the sequence with the highest overall score is generated.

\section{Experimental results}
\label{sec:experiments}
This section presents an overview of the datasets used in the experiments and the preparation steps for model input, that is, images and captions. We will then describe the training and evaluation processes, including the metrics used to train, validate, and evaluate the proposed approach. Finally, we present the experimental setup and settings, then discuss the results.

\subsection{Datasets}
We used the popular benchmark dataset Flickr30K dataset \cite{plummer2015flickr30k} to conduct our experiments. For training, validation, and testing, we used the popular ‘Karpathy’ splits setting \cite{karpathy2015deep} for Flickr30K. Flickr30K dataset has a training set of 29K images, a validation set of 1K images, and a test set of 1K images. 
All images from both datasets are unique and paired with five human-labeled captions.

\subsubsection{Data Preparation}
In order to train the image captioning model on massive datasets such as  Flickr30K, the input data need to be prepared appropriately. To accelerate the training and evaluation process, we preprocessed the images and captions separately into an appropriate format. 

Since we are using a pre-trained encoder (ResNet-101), we preprocess all images to the corresponding format. Pixel values are scaled to $[0,1]$ then normalized using $\mu = [0.485, 0.456, 0.406]$ and $\sigma = [0.229, 0.224, 0.225]$ (mean and standard deviation of the ImageNet images' RGB channels). After that, we resize all images for uniformity. 

We transform captions into a format ready to be fed to the language model. All captions are transformed into a list of tokens with a special start token $ \textlangle$start$\textrangle $ marking the beginning of the sentence, unknown token $ \textlangle$unk$\textrangle $ marking unknown words, pad token $ \textlangle$pad$\textrangle $, and end token $ \textlangle$end$\textrangle $ marking the end of the sentence. All these tokens are converted into integers, where every unique word has an associated integer value in a dictionary file called $word\_map$. We then combine images and their paired captions into a parallel data structure that supports batching and buffering before starting the training, validation, and evaluation. We also save caption lengths before padding since the actual length is essential to avoid wasting time during the decoding process.

\subsection{Model Training}
Image captioning datasets contain a set of images labeled with captions that describe them in natural language. In the supervised training, the loss function is calculated by comparing the predicted caption with the ground truth caption word by word. Then all losses will be summed to obtain a single value, which needs to be minimized.

The cross-entropy loss $L_{entropy}$ is utilized for optimizing the proposed model in the training stage, which minimizes the negative log-likelihood as follows: 
\begin{equation}  
	L_{entropy} = -\log p(S^*|M) = - \sum_{t=2}^{n} \log p(s^*_t|s^*_{1:t-1}),
\end{equation}
where $ M $ is the input image, and $ S^* $ is the ground truth caption.	
The teacher forcing technique \cite{williams1989learning} is used in the training stage to speed up the training process of the recurrent neural network. In this method, the expected output from the training dataset (ground-truth caption) at time step $y(t)$ will be used as an input in the next time step $x(t+1)$ instead of the previously generated word from the language model. This technique is commonly used in sequence-to-sequence generation models in applications such as machine translation, question answering, and image captioning. It solves the problems of instability and slow convergence during the training of recurrent networks that use the previously generated output as an input.

\subsection{Model Evaluation}
After training the model, we can use it to generate new image captions. We use four standard NLP automatic evaluation metrics to assess the quality of the produced captions: Bilingual Evaluation Understudy (BLEU) \cite{papineni2002bleu}, Recall-Oriented Understudy for Gisting Evaluation  (ROUGE) \cite{lin2004rouge}, Metric for Evaluation of Translation with Explicit ORdering (METEOR) \cite{banerjee2005meteor}, and Consensus-based Image Description Evaluation (CIDEr) \cite{vedantam2015cider}. 

We implemented the proposed approach in Python and used Pytorch package as computational framework due to its popularity within the deep learning community and its CUDA parallel computing capabilities. Development and experiments were conducted on an Intel Core i7-9700K CPU with 32GB RAM and an NVIDIA RTX GPU 2060 with 8GB RAM. Google Colab cloud platform was used as the main CUDA platform for the computationally intensive parts of the experiments. 

In the experiments, we evaluated the performance of several models in addition to the proposed ones. Starting from a simple baseline model, we incrementally added additional modules to test their effect on the model's performance. More precisely, we implemented, trained, and compared the following models:	
\begin{itemize}
	\item\textbf{Baseline}: This is the baseline model which has a common encoder-decoder architecture (CNN-RNN). The encoder is ResNet-101, and the decoder uses one layer of LSTM.
	\item\textbf{VisAtt}: This model combines the baseline model with a visual attention-based recurrent module added to the decoder. The attention module applies visual attention to the image features to focus on the salient objects.
	\item\textbf{VisAttRefAtt}: This model combines the baseline model with an attention-based recurrent module and a reflective attention module in the decoder. The reflective attention module attends to the textual information from the language model.
	\item\textbf{RefiningVisAttRefAtt}: This is the proposed architecture shown in Figure \ref{fig:PA2}. It includes a refining module in the encoder and combines an attention-based recurrent module and a reflective attention module in the decoder.
\end{itemize}

\begin{table*} [!t] 
	\centering
	\caption{Description of the tested models.}
	\label{table:desc}
	\footnotesize
	\begin{tabular}{ l p{4cm} l c c c c }
		\toprule
		Model & Description & CNN & Visual attention & Reflective attention & Refining & Global features \\
		\midrule
		Baseline & Baseline encoder-decoder model & ResNet-101 & No & No & No & No \\ 
		\hline
		VisAtt & Baseline + visual attention & ResNet-101 & Yes & No & No & No \\ 
		\hline
		VisAttRefAtt & Baseline + visual attention + reflective attention & ResNet-101 & Yes & Yes & No & No  \\ 
		\hline
		RefiningVisAttRefAtt & Baseline + visual attention + reflective attention + feature refinement  (The proposed architecture shown in Figure \ref{fig:PA2}) & ResNet-101 & Yes & Yes & Yes & Yes\\
		
		\bottomrule
	\end{tabular}
\end{table*}


\subsection{Experimental Settings}

We trained all models with the following settings: the features dimension for ResNet-101 is 2048. The number of heads in the multi-head attention is set to 8.

For each LSTM layer, the hidden and word embedding sizes are set to 1000. The dimension of both attention layers (Visual attention and Reflective attention) is set to 512. The optimizer used is Adamax \cite{kingma2014adam} with a learning rate of 0.002 for the LSTM language model. The number of epochs used for training is 50. We adopt the early stopping method introduced in \cite{xu2015show}, which uses BLEU evaluation metric to evaluate the model's performance on the validation set. Therefore, the model will stop the training if the value of BLEU score does not improve for 12 consecutive epochs; otherwise, the model will be trained for up to 50 epochs. Weight normalization is used o prevent the model from over-fitting. The dropout value is set to 0.5, and the batch size is set to 64. Finally, the beam search size is set to 5.

\subsection{Result and Analysis}

All experimental results are summarized in Table~\ref{table:2}. The higher the value in each column, the better. We reported BLEU with four different N-gram lengths: N=1, N=2, N=3, and N=4.

\begin{table*} [!t] 
	\centering
	\caption{Models performance on the Flickr30K Karpathy test split.}
	\label{table:2}
	\footnotesize
	\begin{tabular}{ l r r r r r r r }
		\toprule
		Model & BLEU-1 & BLEU-2 & BLEU-3 & BLEU-4 & METEOR & ROUGE-L & CIDEr \\
		\midrule
		Baseline & 62.82 & 43.49 & 29.69 & 20.07 & 17.94 & 43.62 & 40.81 \\ 
		\hline
		VisAtt & 67.34 & 48.49 & 34.60 & 24.22 & 19.77 & 46.12 & 51.08 \\
		\hline
		VisAttRefAtt & 67.39 & 48.69 & 35.16 & 24.91 & 20.05 & 46.48 & 52.36 \\ 
		\hline
		RefiningVisAttRefAtt & 67.36 & 48.54 & 34.81 & 24.86 & 20.30 & 46.51 & 52.69 \\ 
		\bottomrule
	\end{tabular}
\end{table*}

Table~\ref{table:2} shows the performance results on the Flickr30K dataset. First, as we can see, the baseline model has the lowest performance, which is expected since it only uses CNN-RNN without any modifications or additions. The visual attention module (VisAtt) improves the performance of the baseline model in all evaluation metrics (BLUEs, METEOR, ROUGE-L, and CIDEr). Upon adding the reflective attention module (VisAttRefAtt model), the performance improves further compared to using only visual attention. When we add the refinement module (RefiningVisAttRefAtt model ), we see a slight degradation in the BLEU score. However, the other evaluation metrics (METEOR, ROUGE-L, and CIDEr) show improvement.

\begin{figure} [!t] 
	\centering 
	\includegraphics[width=1\linewidth]{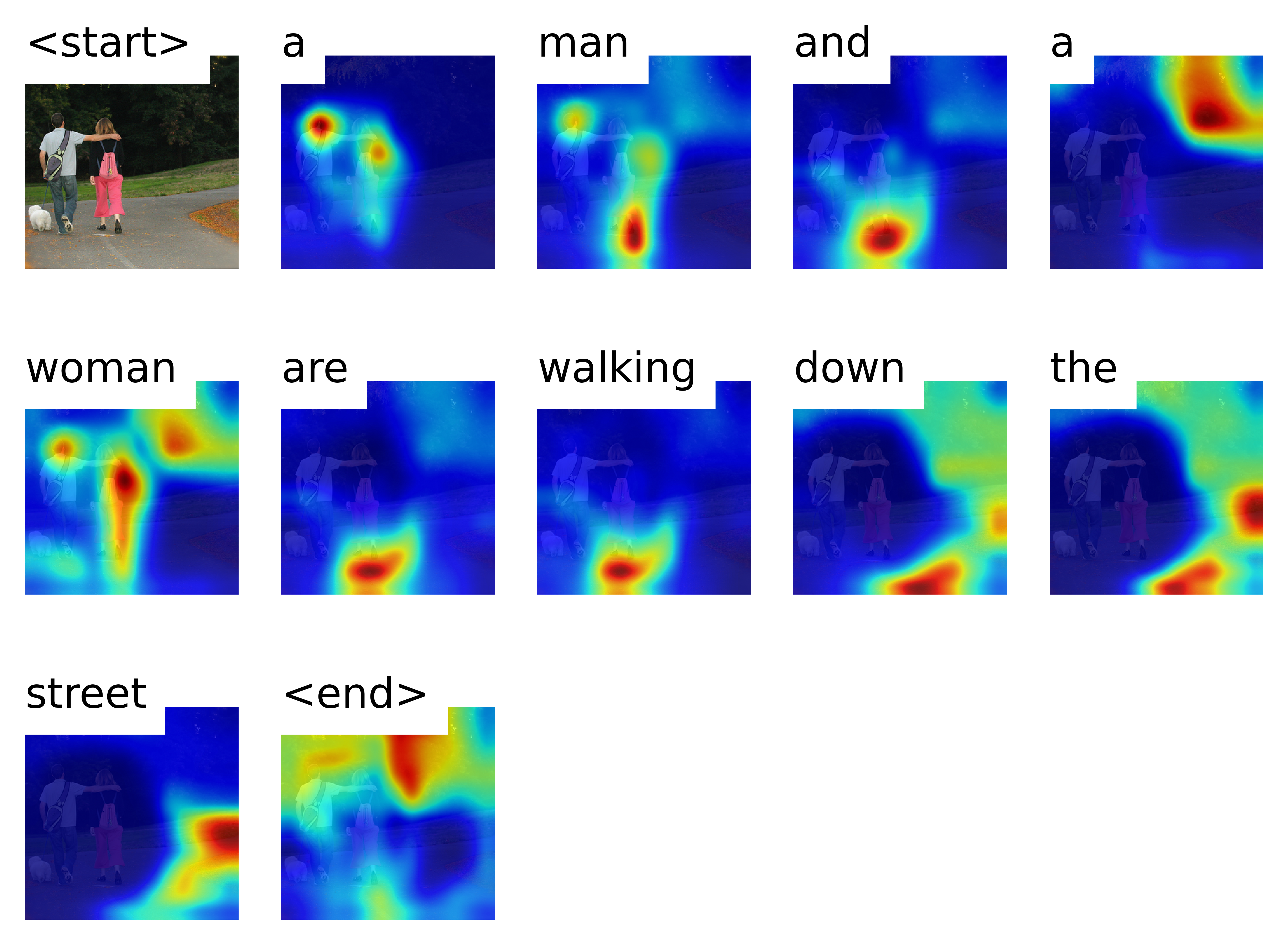}
	\caption{Visualization of attention weights learned by RefiningVisAttRefAtt model}
	\label{fig:AVR}
\end{figure}

Figure~\ref{fig:AVR} shows the model attention when using feature refinement on a sample image from the Flickr30K test split. We visualize the attention weights used to generate each word in the caption. Each picture shows the generated word and the corersponding attention weights $\alpha_t$ at each time step $t$.	

\section{Conclusion}
\label{sec:conclusion}
Despite the significant progress made in the field during the last decade, it is clear from the literature that image captioning is still an active research area. Numerous approaches have been proposed recently to improve the performance of captioning systems and address existing issues. The challenges currently tackled by researchers include generating better quality captions, developing more sophisticated and powerful visual and language network architectures, and improving image captioning datasets and evaluation metrics.

In this line of research, we propose an encoder-decoder-based image captioning system with an encoder that extracts the spatial and global features and uses a refining model that models the relationships among objects in the target image. The decoder includes attention-based recurrent and reflective attention modules for better quality caption generation. The proposed approach takes the relationship between the target image objects and focuses on both the visual and the textual information by applying different attention mechanisms. In the experimental evaluation, we developed, trained, evaluated, and compared several models, including the proposed one. The experimental results show the effectiveness of the proposed approach.	

As future work, we propose to focus more on the encoder side and try to learn more from the visual features by applying  Faster R-CNN to learn the relationships between the different objects in the image.

\backmatter

\bmhead{Acknowledgments}

This research work is supported by the Research Center, CCIS, King Saud University, Riyadh, Saudi Arabia.



\begin{thebibliography}{42}
	\ifx \bisbn   \undefined \def \bisbn  #1{ISBN #1}\fi
	\ifx \binits  \undefined \def \binits#1{#1}\fi
	\ifx \bauthor  \undefined \def \bauthor#1{#1}\fi
	\ifx \batitle  \undefined \def \batitle#1{#1}\fi
	\ifx \bjtitle  \undefined \def \bjtitle#1{#1}\fi
	\ifx \bvolume  \undefined \def \bvolume#1{\textbf{#1}}\fi
	\ifx \byear  \undefined \def \byear#1{#1}\fi
	\ifx \bissue  \undefined \def \bissue#1{#1}\fi
	\ifx \bfpage  \undefined \def \bfpage#1{#1}\fi
	\ifx \blpage  \undefined \def \blpage #1{#1}\fi
	\ifx \burl  \undefined \def \burl#1{\textsf{#1}}\fi
	\ifx \doiurl  \undefined \def \doiurl#1{\url{https://doi.org/#1}}\fi
	\ifx \betal  \undefined \def \betal{\textit{et al.}}\fi
	\ifx \binstitute  \undefined \def \binstitute#1{#1}\fi
	\ifx \binstitutionaled  \undefined \def \binstitutionaled#1{#1}\fi
	\ifx \bctitle  \undefined \def \bctitle#1{#1}\fi
	\ifx \beditor  \undefined \def \beditor#1{#1}\fi
	\ifx \bpublisher  \undefined \def \bpublisher#1{#1}\fi
	\ifx \bbtitle  \undefined \def \bbtitle#1{#1}\fi
	\ifx \bedition  \undefined \def \bedition#1{#1}\fi
	\ifx \bseriesno  \undefined \def \bseriesno#1{#1}\fi
	\ifx \blocation  \undefined \def \blocation#1{#1}\fi
	\ifx \bsertitle  \undefined \def \bsertitle#1{#1}\fi
	\ifx \bsnm \undefined \def \bsnm#1{#1}\fi
	\ifx \bsuffix \undefined \def \bsuffix#1{#1}\fi
	\ifx \bparticle \undefined \def \bparticle#1{#1}\fi
	\ifx \barticle \undefined \def \barticle#1{#1}\fi
	\bibcommenthead
	\ifx \bconfdate \undefined \def \bconfdate #1{#1}\fi
	\ifx \botherref \undefined \def \botherref #1{#1}\fi
	\ifx \url \undefined \def \url#1{\textsf{#1}}\fi
	\ifx \bchapter \undefined \def \bchapter#1{#1}\fi
	\ifx \bbook \undefined \def \bbook#1{#1}\fi
	\ifx \bcomment \undefined \def \bcomment#1{#1}\fi
	\ifx \oauthor \undefined \def \oauthor#1{#1}\fi
	\ifx \citeauthoryear \undefined \def \citeauthoryear#1{#1}\fi
	\ifx \endbibitem  \undefined \def \endbibitem {}\fi
	\ifx \bconflocation  \undefined \def \bconflocation#1{#1}\fi
	\ifx \arxivurl  \undefined \def \arxivurl#1{\textsf{#1}}\fi
	\csname PreBibitemsHook\endcsname
	
	\bibitem{torralba2003context}
	\begin{botherref}
		\oauthor{\bsnm{Torralba}, \binits{A.}},
		\oauthor{\bsnm{Murphy}, \binits{K.P.}},
		\oauthor{\bsnm{Freeman}, \binits{W.T.}},
		\oauthor{\bsnm{Rubin}, \binits{M.A.}}:
		Context-based vision system for place and object recognition.
		Proceedings Ninth IEEE International Conference on Computer Vision,
		273--2801
		(2003)
	\end{botherref}
	\endbibitem
	
	\bibitem{vinyals2016show}
	\begin{barticle}
		\bauthor{\bsnm{Vinyals}, \binits{O.}},
		\bauthor{\bsnm{Toshev}, \binits{A.}},
		\bauthor{\bsnm{Bengio}, \binits{S.}},
		\bauthor{\bsnm{Erhan}, \binits{D.}}:
		\batitle{Show and tell: Lessons learned from the 2015 mscoco image captioning
			challenge}.
		\bjtitle{IEEE transactions on pattern analysis and machine intelligence}
		\bvolume{39}(\bissue{4}),
		\bfpage{652}--\blpage{663}
		(\byear{2016})
	\end{barticle}
	\endbibitem
	
	\bibitem{huang2019attention}
	\begin{bchapter}
		\bauthor{\bsnm{Huang}, \binits{L.}},
		\bauthor{\bsnm{Wang}, \binits{W.}},
		\bauthor{\bsnm{Chen}, \binits{J.}},
		\bauthor{\bsnm{Wei}, \binits{X.-Y.}}:
		\bctitle{Attention on attention for image captioning}.
		In: \bbtitle{Proceedings of the IEEE International Conference on Computer
			Vision},
		pp. \bfpage{4634}--\blpage{4643}
		(\byear{2019})
	\end{bchapter}
	\endbibitem
	
	\bibitem{ke2019reflective}
	\begin{bchapter}
		\bauthor{\bsnm{Ke}, \binits{L.}},
		\bauthor{\bsnm{Pei}, \binits{W.}},
		\bauthor{\bsnm{Li}, \binits{R.}},
		\bauthor{\bsnm{Shen}, \binits{X.}},
		\bauthor{\bsnm{Tai}, \binits{Y.-W.}}:
		\bctitle{Reflective decoding network for image captioning}.
		In: \bbtitle{Proceedings of the IEEE International Conference on Computer
			Vision},
		pp. \bfpage{8888}--\blpage{8897}
		(\byear{2019})
	\end{bchapter}
	\endbibitem
	
	\bibitem{hossain2019comprehensive}
	\begin{barticle}
		\bauthor{\bsnm{Hossain}, \binits{M.}},
		\bauthor{\bsnm{Sohel}, \binits{F.}},
		\bauthor{\bsnm{Shiratuddin}, \binits{M.F.}},
		\bauthor{\bsnm{Laga}, \binits{H.}}:
		\batitle{A comprehensive survey of deep learning for image captioning}.
		\bjtitle{ACM Computing Surveys (CSUR)}
		\bvolume{51}(\bissue{6}),
		\bfpage{118}
		(\byear{2019})
	\end{barticle}
	\endbibitem
	
	\bibitem{bai2018survey}
	\begin{barticle}
		\bauthor{\bsnm{Bai}, \binits{S.}},
		\bauthor{\bsnm{An}, \binits{S.}}:
		\batitle{A survey on automatic image caption generation}.
		\bjtitle{Neurocomputing}
		\bvolume{311},
		\bfpage{291}--\blpage{304}
		(\byear{2018})
	\end{barticle}
	\endbibitem
	
	\bibitem{farhadi2010every}
	\begin{bchapter}
		\bauthor{\bsnm{Farhadi}, \binits{A.}},
		\bauthor{\bsnm{Hejrati}, \binits{M.}},
		\bauthor{\bsnm{Sadeghi}, \binits{M.A.}},
		\bauthor{\bsnm{Young}, \binits{P.}},
		\bauthor{\bsnm{Rashtchian}, \binits{C.}},
		\bauthor{\bsnm{Hockenmaier}, \binits{J.}},
		\bauthor{\bsnm{Forsyth}, \binits{D.}}:
		\bctitle{Every picture tells a story: Generating sentences from images}.
		In: \bbtitle{European Conference on Computer Vision},
		pp. \bfpage{15}--\blpage{29}
		(\byear{2010}).
		\bcomment{Springer}
	\end{bchapter}
	\endbibitem
	
	\bibitem{hodosh2013framing}
	\begin{barticle}
		\bauthor{\bsnm{Hodosh}, \binits{M.}},
		\bauthor{\bsnm{Young}, \binits{P.}},
		\bauthor{\bsnm{Hockenmaier}, \binits{J.}}:
		\batitle{Framing image description as a ranking task: Data, models and
			evaluation metrics}.
		\bjtitle{Journal of Artificial Intelligence Research}
		\bvolume{47},
		\bfpage{853}--\blpage{899}
		(\byear{2013})
	\end{barticle}
	\endbibitem
	
	\bibitem{lin2014microsoft}
	\begin{bchapter}
		\bauthor{\bsnm{Lin}, \binits{T.-Y.}},
		\bauthor{\bsnm{Maire}, \binits{M.}},
		\bauthor{\bsnm{Belongie}, \binits{S.}},
		\bauthor{\bsnm{Hays}, \binits{J.}},
		\bauthor{\bsnm{Perona}, \binits{P.}},
		\bauthor{\bsnm{Ramanan}, \binits{D.}},
		\bauthor{\bsnm{Doll{\'a}r}, \binits{P.}},
		\bauthor{\bsnm{Zitnick}, \binits{C.L.}}:
		\bctitle{Microsoft coco: Common objects in context}.
		In: \bbtitle{European Conference on Computer Vision},
		pp. \bfpage{740}--\blpage{755}
		(\byear{2014}).
		\bcomment{Springer}
	\end{bchapter}
	\endbibitem
	
	\bibitem{sharma2018conceptual}
	\begin{bchapter}
		\bauthor{\bsnm{Sharma}, \binits{P.}},
		\bauthor{\bsnm{Ding}, \binits{N.}},
		\bauthor{\bsnm{Goodman}, \binits{S.}},
		\bauthor{\bsnm{Soricut}, \binits{R.}}:
		\bctitle{Conceptual captions: A cleaned, hypernymed, image alt-text dataset for
			automatic image captioning}.
		In: \bbtitle{Proceedings of the 56th Annual Meeting of the Association for
			Computational Linguistics (Volume 1: Long Papers)},
		pp. \bfpage{2556}--\blpage{2565}
		(\byear{2018})
	\end{bchapter}
	\endbibitem
	
	\bibitem{yang2011corpus}
	\begin{bchapter}
		\bauthor{\bsnm{Yang}, \binits{Y.}},
		\bauthor{\bsnm{Teo}, \binits{C.L.}},
		\bauthor{\bsnm{Daum{\'e}~III}, \binits{H.}},
		\bauthor{\bsnm{Aloimonos}, \binits{Y.}}:
		\bctitle{Corpus-guided sentence generation of natural images}.
		In: \bbtitle{Proceedings of the Conference on Empirical Methods in Natural
			Language Processing},
		pp. \bfpage{444}--\blpage{454}
		(\byear{2011}).
		\bcomment{Association for Computational Linguistics}
	\end{bchapter}
	\endbibitem
	
	\bibitem{mitchell2012midge}
	\begin{bchapter}
		\bauthor{\bsnm{Mitchell}, \binits{M.}},
		\bauthor{\bsnm{Han}, \binits{X.}},
		\bauthor{\bsnm{Dodge}, \binits{J.}},
		\bauthor{\bsnm{Mensch}, \binits{A.}},
		\bauthor{\bsnm{Goyal}, \binits{A.}},
		\bauthor{\bsnm{Berg}, \binits{A.}},
		\bauthor{\bsnm{Yamaguchi}, \binits{K.}},
		\bauthor{\bsnm{Berg}, \binits{T.}},
		\bauthor{\bsnm{Stratos}, \binits{K.}},
		\bauthor{\bsnm{Daum{\'e}~III}, \binits{H.}}:
		\bctitle{Midge: Generating image descriptions from computer vision detections}.
		In: \bbtitle{Proceedings of the 13th Conference of the European Chapter of the
			Association for Computational Linguistics},
		pp. \bfpage{747}--\blpage{756}
		(\byear{2012}).
		\bcomment{Association for Computational Linguistics}
	\end{bchapter}
	\endbibitem
	
	\bibitem{kiros2014multimodal}
	\begin{bchapter}
		\bauthor{\bsnm{Kiros}, \binits{R.}},
		\bauthor{\bsnm{Salakhutdinov}, \binits{R.}},
		\bauthor{\bsnm{Zemel}, \binits{R.}}:
		\bctitle{Multimodal neural language models}.
		In: \bbtitle{International Conference on Machine Learning},
		pp. \bfpage{595}--\blpage{603}
		(\byear{2014})
	\end{bchapter}
	\endbibitem
	
	\bibitem{lecun1998gradient}
	\begin{barticle}
		\bauthor{\bsnm{LeCun}, \binits{Y.}},
		\bauthor{\bsnm{Bottou}, \binits{L.}},
		\bauthor{\bsnm{Bengio}, \binits{Y.}},
		\bauthor{\bsnm{Haffner}, \binits{P.}}, \betal:
		\batitle{Gradient-based learning applied to document recognition}.
		\bjtitle{Proceedings of the IEEE}
		\bvolume{86}(\bissue{11}),
		\bfpage{2278}--\blpage{2324}
		(\byear{1998})
	\end{barticle}
	\endbibitem
	
	\bibitem{kiros2014unifying}
	\begin{botherref}
		\oauthor{\bsnm{Kiros}, \binits{R.}},
		\oauthor{\bsnm{Salakhutdinov}, \binits{R.}},
		\oauthor{\bsnm{Zemel}, \binits{R.S.}}:
		Unifying visual-semantic embeddings with multimodal neural language models.
		arXiv preprint arXiv:1411.2539
		(2014)
	\end{botherref}
	\endbibitem
	
	\bibitem{vinyals2015show}
	\begin{bchapter}
		\bauthor{\bsnm{Vinyals}, \binits{O.}},
		\bauthor{\bsnm{Toshev}, \binits{A.}},
		\bauthor{\bsnm{Bengio}, \binits{S.}},
		\bauthor{\bsnm{Erhan}, \binits{D.}}:
		\bctitle{Show and tell: A neural image caption generator}.
		In: \bbtitle{Proceedings of the IEEE Conference on Computer Vision and Pattern
			Recognition},
		pp. \bfpage{3156}--\blpage{3164}
		(\byear{2015})
	\end{bchapter}
	\endbibitem
	
	\bibitem{fang2015captions}
	\begin{bchapter}
		\bauthor{\bsnm{Fang}, \binits{H.}},
		\bauthor{\bsnm{Gupta}, \binits{S.}},
		\bauthor{\bsnm{Iandola}, \binits{F.}},
		\bauthor{\bsnm{Srivastava}, \binits{R.K.}},
		\bauthor{\bsnm{Deng}, \binits{L.}},
		\bauthor{\bsnm{Doll{\'a}r}, \binits{P.}},
		\bauthor{\bsnm{Gao}, \binits{J.}},
		\bauthor{\bsnm{He}, \binits{X.}},
		\bauthor{\bsnm{Mitchell}, \binits{M.}},
		\bauthor{\bsnm{Platt}, \binits{J.C.}}, \betal:
		\bctitle{From captions to visual concepts and back}.
		In: \bbtitle{Proceedings of the IEEE Conference on Computer Vision and Pattern
			Recognition},
		pp. \bfpage{1473}--\blpage{1482}
		(\byear{2015})
	\end{bchapter}
	\endbibitem
	
	\bibitem{tran2016rich}
	\begin{bchapter}
		\bauthor{\bsnm{Tran}, \binits{K.}},
		\bauthor{\bsnm{He}, \binits{X.}},
		\bauthor{\bsnm{Zhang}, \binits{L.}},
		\bauthor{\bsnm{Sun}, \binits{J.}},
		\bauthor{\bsnm{Carapcea}, \binits{C.}},
		\bauthor{\bsnm{Thrasher}, \binits{C.}},
		\bauthor{\bsnm{Buehler}, \binits{C.}},
		\bauthor{\bsnm{Sienkiewicz}, \binits{C.}}:
		\bctitle{Rich image captioning in the wild}.
		In: \bbtitle{Proceedings of the IEEE Conference on Computer Vision and Pattern
			Recognition Workshops},
		pp. \bfpage{49}--\blpage{56}
		(\byear{2016})
	\end{bchapter}
	\endbibitem
	
	\bibitem{xu2015show}
	\begin{bchapter}
		\bauthor{\bsnm{Xu}, \binits{K.}},
		\bauthor{\bsnm{Ba}, \binits{J.}},
		\bauthor{\bsnm{Kiros}, \binits{R.}},
		\bauthor{\bsnm{Cho}, \binits{K.}},
		\bauthor{\bsnm{Courville}, \binits{A.}},
		\bauthor{\bsnm{Salakhudinov}, \binits{R.}},
		\bauthor{\bsnm{Zemel}, \binits{R.}},
		\bauthor{\bsnm{Bengio}, \binits{Y.}}:
		\bctitle{Show, attend and tell: Neural image caption generation with visual
			attention}.
		In: \bbtitle{International Conference on Machine Learning},
		pp. \bfpage{2048}--\blpage{2057}
		(\byear{2015})
	\end{bchapter}
	\endbibitem
	
	\bibitem{lu2017knowing}
	\begin{bchapter}
		\bauthor{\bsnm{Lu}, \binits{J.}},
		\bauthor{\bsnm{Xiong}, \binits{C.}},
		\bauthor{\bsnm{Parikh}, \binits{D.}},
		\bauthor{\bsnm{Socher}, \binits{R.}}:
		\bctitle{Knowing when to look: Adaptive attention via a visual sentinel for
			image captioning}.
		In: \bbtitle{Proceedings of the IEEE Conference on Computer Vision and Pattern
			Recognition},
		pp. \bfpage{375}--\blpage{383}
		(\byear{2017})
	\end{bchapter}
	\endbibitem
	
	\bibitem{pedersoli2017areas}
	\begin{bchapter}
		\bauthor{\bsnm{Pedersoli}, \binits{M.}},
		\bauthor{\bsnm{Lucas}, \binits{T.}},
		\bauthor{\bsnm{Schmid}, \binits{C.}},
		\bauthor{\bsnm{Verbeek}, \binits{J.}}:
		\bctitle{Areas of attention for image captioning}.
		In: \bbtitle{Proceedings of the IEEE International Conference on Computer
			Vision},
		pp. \bfpage{1242}--\blpage{1250}
		(\byear{2017})
	\end{bchapter}
	\endbibitem
	
	\bibitem{donahue2015long}
	\begin{bchapter}
		\bauthor{\bsnm{Donahue}, \binits{J.}},
		\bauthor{\bsnm{Anne~Hendricks}, \binits{L.}},
		\bauthor{\bsnm{Guadarrama}, \binits{S.}},
		\bauthor{\bsnm{Rohrbach}, \binits{M.}},
		\bauthor{\bsnm{Venugopalan}, \binits{S.}},
		\bauthor{\bsnm{Saenko}, \binits{K.}},
		\bauthor{\bsnm{Darrell}, \binits{T.}}:
		\bctitle{Long-term recurrent convolutional networks for visual recognition and
			description}.
		In: \bbtitle{Proceedings of the IEEE Conference on Computer Vision and Pattern
			Recognition},
		pp. \bfpage{2625}--\blpage{2634}
		(\byear{2015})
	\end{bchapter}
	\endbibitem
	
	\bibitem{karpathy2015deep}
	\begin{bchapter}
		\bauthor{\bsnm{Karpathy}, \binits{A.}},
		\bauthor{\bsnm{Fei-Fei}, \binits{L.}}:
		\bctitle{Deep visual-semantic alignments for generating image descriptions}.
		In: \bbtitle{Proceedings of the IEEE Conference on Computer Vision and Pattern
			Recognition},
		pp. \bfpage{3128}--\blpage{3137}
		(\byear{2015})
	\end{bchapter}
	\endbibitem
	
	\bibitem{kulkarni2013babytalk}
	\begin{barticle}
		\bauthor{\bsnm{Kulkarni}, \binits{G.}},
		\bauthor{\bsnm{Premraj}, \binits{V.}},
		\bauthor{\bsnm{Ordonez}, \binits{V.}},
		\bauthor{\bsnm{Dhar}, \binits{S.}},
		\bauthor{\bsnm{Li}, \binits{S.}},
		\bauthor{\bsnm{Choi}, \binits{Y.}},
		\bauthor{\bsnm{Berg}, \binits{A.C.}},
		\bauthor{\bsnm{Berg}, \binits{T.L.}}:
		\batitle{Babytalk: Understanding and generating simple image descriptions}.
		\bjtitle{IEEE Transactions on Pattern Analysis and Machine Intelligence}
		\bvolume{35}(\bissue{12}),
		\bfpage{2891}--\blpage{2903}
		(\byear{2013})
	\end{barticle}
	\endbibitem
	
	\bibitem{you2016image}
	\begin{bchapter}
		\bauthor{\bsnm{You}, \binits{Q.}},
		\bauthor{\bsnm{Jin}, \binits{H.}},
		\bauthor{\bsnm{Wang}, \binits{Z.}},
		\bauthor{\bsnm{Fang}, \binits{C.}},
		\bauthor{\bsnm{Luo}, \binits{J.}}:
		\bctitle{Image captioning with semantic attention}.
		In: \bbtitle{Proceedings of the IEEE Conference on Computer Vision and Pattern
			Recognition},
		pp. \bfpage{4651}--\blpage{4659}
		(\byear{2016})
	\end{bchapter}
	\endbibitem
	
	\bibitem{wu2017image}
	\begin{barticle}
		\bauthor{\bsnm{Wu}, \binits{Q.}},
		\bauthor{\bsnm{Shen}, \binits{C.}},
		\bauthor{\bsnm{Wang}, \binits{P.}},
		\bauthor{\bsnm{Dick}, \binits{A.}},
		\bauthor{\bparticle{van~den} \bsnm{Hengel}, \binits{A.}}:
		\batitle{Image captioning and visual question answering based on attributes and
			external knowledge}.
		\bjtitle{IEEE transactions on pattern analysis and machine intelligence}
		\bvolume{40}(\bissue{6}),
		\bfpage{1367}--\blpage{1381}
		(\byear{2017})
	\end{barticle}
	\endbibitem
	
	\bibitem{simonyan2014very}
	\begin{botherref}
		\oauthor{\bsnm{Simonyan}, \binits{K.}},
		\oauthor{\bsnm{Zisserman}, \binits{A.}}:
		Very deep convolutional networks for large-scale image recognition.
		arXiv preprint arXiv:1409.1556
		(2014)
	\end{botherref}
	\endbibitem
	
	\bibitem{he2016deep}
	\begin{bchapter}
		\bauthor{\bsnm{He}, \binits{K.}},
		\bauthor{\bsnm{Zhang}, \binits{X.}},
		\bauthor{\bsnm{Ren}, \binits{S.}},
		\bauthor{\bsnm{Sun}, \binits{J.}}:
		\bctitle{Deep residual learning for image recognition}.
		In: \bbtitle{Proceedings of the IEEE Conference on Computer Vision and Pattern
			Recognition},
		pp. \bfpage{770}--\blpage{778}
		(\byear{2016})
	\end{bchapter}
	\endbibitem
	
	\bibitem{krizhevsky2012imagenet}
	\begin{bchapter}
		\bauthor{\bsnm{Krizhevsky}, \binits{A.}},
		\bauthor{\bsnm{Sutskever}, \binits{I.}},
		\bauthor{\bsnm{Hinton}, \binits{G.E.}}:
		\bctitle{Imagenet classification with deep convolutional neural networks}.
		In: \bbtitle{Advances in Neural Information Processing Systems},
		pp. \bfpage{1097}--\blpage{1105}
		(\byear{2012})
	\end{bchapter}
	\endbibitem
	
	\bibitem{hochreiter1997long}
	\begin{barticle}
		\bauthor{\bsnm{Hochreiter}, \binits{S.}},
		\bauthor{\bsnm{Schmidhuber}, \binits{J.}}:
		\batitle{Long short-term memory}.
		\bjtitle{Neural computation}
		\bvolume{9}(\bissue{8}),
		\bfpage{1735}--\blpage{1780}
		(\byear{1997})
	\end{barticle}
	\endbibitem
	
	\bibitem{rumelhart1986learning}
	\begin{barticle}
		\bauthor{\bsnm{Rumelhart}, \binits{D.E.}},
		\bauthor{\bsnm{Hinton}, \binits{G.E.}},
		\bauthor{\bsnm{Williams}, \binits{R.J.}}:
		\batitle{Learning representations by back-propagating errors}.
		\bjtitle{nature}
		\bvolume{323}(\bissue{6088}),
		\bfpage{533}--\blpage{536}
		(\byear{1986})
	\end{barticle}
	\endbibitem
	
	\bibitem{anderson2018bottom}
	\begin{bchapter}
		\bauthor{\bsnm{Anderson}, \binits{P.}},
		\bauthor{\bsnm{He}, \binits{X.}},
		\bauthor{\bsnm{Buehler}, \binits{C.}},
		\bauthor{\bsnm{Teney}, \binits{D.}},
		\bauthor{\bsnm{Johnson}, \binits{M.}},
		\bauthor{\bsnm{Gould}, \binits{S.}},
		\bauthor{\bsnm{Zhang}, \binits{L.}}:
		\bctitle{Bottom-up and top-down attention for image captioning and visual
			question answering}.
		In: \bbtitle{Proceedings of the IEEE Conference on Computer Vision and Pattern
			Recognition},
		pp. \bfpage{6077}--\blpage{6086}
		(\byear{2018})
	\end{bchapter}
	\endbibitem
	
	\bibitem{russakovsky2015imagenet}
	\begin{barticle}
		\bauthor{\bsnm{Russakovsky}, \binits{O.}},
		\bauthor{\bsnm{Deng}, \binits{J.}},
		\bauthor{\bsnm{Su}, \binits{H.}},
		\bauthor{\bsnm{Krause}, \binits{J.}},
		\bauthor{\bsnm{Satheesh}, \binits{S.}},
		\bauthor{\bsnm{Ma}, \binits{S.}},
		\bauthor{\bsnm{Huang}, \binits{Z.}},
		\bauthor{\bsnm{Karpathy}, \binits{A.}},
		\bauthor{\bsnm{Khosla}, \binits{A.}},
		\bauthor{\bsnm{Bernstein}, \binits{M.}}, \betal:
		\batitle{Imagenet large scale visual recognition challenge}.
		\bjtitle{International journal of computer vision}
		\bvolume{115}(\bissue{3}),
		\bfpage{211}--\blpage{252}
		(\byear{2015})
	\end{barticle}
	\endbibitem
	
	\bibitem{vaswani2017attention}
	\begin{bchapter}
		\bauthor{\bsnm{Vaswani}, \binits{A.}},
		\bauthor{\bsnm{Shazeer}, \binits{N.}},
		\bauthor{\bsnm{Parmar}, \binits{N.}},
		\bauthor{\bsnm{Uszkoreit}, \binits{J.}},
		\bauthor{\bsnm{Jones}, \binits{L.}},
		\bauthor{\bsnm{Gomez}, \binits{A.N.}},
		\bauthor{\bsnm{Kaiser}, \binits{{\L}.}},
		\bauthor{\bsnm{Polosukhin}, \binits{I.}}:
		\bctitle{Attention is all you need}.
		In: \bbtitle{Advances in Neural Information Processing Systems},
		pp. \bfpage{5998}--\blpage{6008}
		(\byear{2017})
	\end{bchapter}
	\endbibitem
	
	\bibitem{ba2016layer}
	\begin{botherref}
		\oauthor{\bsnm{Ba}, \binits{J.L.}},
		\oauthor{\bsnm{Kiros}, \binits{J.R.}},
		\oauthor{\bsnm{Hinton}, \binits{G.E.}}:
		Layer normalization.
		arXiv preprint arXiv:1607.06450
		(2016)
	\end{botherref}
	\endbibitem
	
	\bibitem{plummer2015flickr30k}
	\begin{bchapter}
		\bauthor{\bsnm{Plummer}, \binits{B.A.}},
		\bauthor{\bsnm{Wang}, \binits{L.}},
		\bauthor{\bsnm{Cervantes}, \binits{C.M.}},
		\bauthor{\bsnm{Caicedo}, \binits{J.C.}},
		\bauthor{\bsnm{Hockenmaier}, \binits{J.}},
		\bauthor{\bsnm{Lazebnik}, \binits{S.}}:
		\bctitle{Flickr30k entities: Collecting region-to-phrase correspondences for
			richer image-to-sentence models}.
		In: \bbtitle{Proceedings of the IEEE International Conference on Computer
			Vision},
		pp. \bfpage{2641}--\blpage{2649}
		(\byear{2015})
	\end{bchapter}
	\endbibitem
	
	\bibitem{williams1989learning}
	\begin{barticle}
		\bauthor{\bsnm{Williams}, \binits{R.J.}},
		\bauthor{\bsnm{Zipser}, \binits{D.}}:
		\batitle{A learning algorithm for continually running fully recurrent neural
			networks}.
		\bjtitle{Neural computation}
		\bvolume{1}(\bissue{2}),
		\bfpage{270}--\blpage{280}
		(\byear{1989})
	\end{barticle}
	\endbibitem
	
	\bibitem{papineni2002bleu}
	\begin{bchapter}
		\bauthor{\bsnm{Papineni}, \binits{K.}},
		\bauthor{\bsnm{Roukos}, \binits{S.}},
		\bauthor{\bsnm{Ward}, \binits{T.}},
		\bauthor{\bsnm{Zhu}, \binits{W.-J.}}:
		\bctitle{Bleu: a method for automatic evaluation of machine translation}.
		In: \bbtitle{Proceedings of the 40th Annual Meeting on Association for
			Computational Linguistics},
		pp. \bfpage{311}--\blpage{318}
		(\byear{2002}).
		\bcomment{Association for Computational Linguistics}
	\end{bchapter}
	\endbibitem
	
	\bibitem{lin2004rouge}
	\begin{bchapter}
		\bauthor{\bsnm{Lin}, \binits{C.-Y.}}:
		\bctitle{Rouge: A package for automatic evaluation of summaries}.
		In: \bbtitle{Text Summarization Branches Out},
		pp. \bfpage{74}--\blpage{81}
		(\byear{2004})
	\end{bchapter}
	\endbibitem
	
	\bibitem{banerjee2005meteor}
	\begin{bchapter}
		\bauthor{\bsnm{Banerjee}, \binits{S.}},
		\bauthor{\bsnm{Lavie}, \binits{A.}}:
		\bctitle{Meteor: An automatic metric for mt evaluation with improved
			correlation with human judgments}.
		In: \bbtitle{Proceedings of the Acl Workshop on Intrinsic and Extrinsic
			Evaluation Measures for Machine Translation And/or Summarization},
		pp. \bfpage{65}--\blpage{72}
		(\byear{2005})
	\end{bchapter}
	\endbibitem
	
	\bibitem{vedantam2015cider}
	\begin{bchapter}
		\bauthor{\bsnm{Vedantam}, \binits{R.}},
		\bauthor{\bsnm{Lawrence~Zitnick}, \binits{C.}},
		\bauthor{\bsnm{Parikh}, \binits{D.}}:
		\bctitle{Cider: Consensus-based image description evaluation}.
		In: \bbtitle{Proceedings of the IEEE Conference on Computer Vision and Pattern
			Recognition},
		pp. \bfpage{4566}--\blpage{4575}
		(\byear{2015})
	\end{bchapter}
	\endbibitem
	
	\bibitem{kingma2014adam}
	\begin{botherref}
		\oauthor{\bsnm{Kingma}, \binits{D.P.}},
		\oauthor{\bsnm{Ba}, \binits{J.}}:
		Adam: A method for stochastic optimization.
		arXiv preprint arXiv:1412.6980
		(2014)
	\end{botherref}
	\endbibitem
	
\end{thebibliography}



\end{document}